\documentclass[10pt,twocolumn,letterpaper]{article}

\usepackage[pagenumbers]{cvpr} % To force page numbers, e.g. for an arXiv version

\usepackage{graphicx}
\usepackage{amsmath}
\usepackage{amssymb}
\usepackage{booktabs}

\usepackage[pagebackref,breaklinks,colorlinks]{hyperref}

\usepackage[capitalize]{cleveref}
\crefname{section}{Sec.}{Secs.}
\Crefname{section}{Section}{Sections}
\Crefname{table}{Table}{Tables}
\crefname{table}{Tab.}{Tabs.}

\def\cvprPaperID{*****} % *** Enter the CVPR Paper ID here
\def\confName{CVPR}
\def\confYear{2022}

\begin{document}

\title{TrackNet: A Triplet metric-based method for Multi-Target Multi-Camera Vehicle Tracking}
\author{David Serrano Lozano, Francesc Net Barnés, \\
Juan Antonio Rodríguez García and Igor Ugarte Molinet\\\\
Universitat Politecnica de Catalunya\\
Barcelona, Spain\\
\tt\small  99d.serrano@gmail.com, \tt\small francescnet@gmail.com, \\
\tt\small juanantonio.rodriguez@upf.edu,  \tt\small igorugarte.cvm@gmail.com
}

\maketitle

\begin{figure*}
\begin{center}
\includegraphics[scale=0.35]{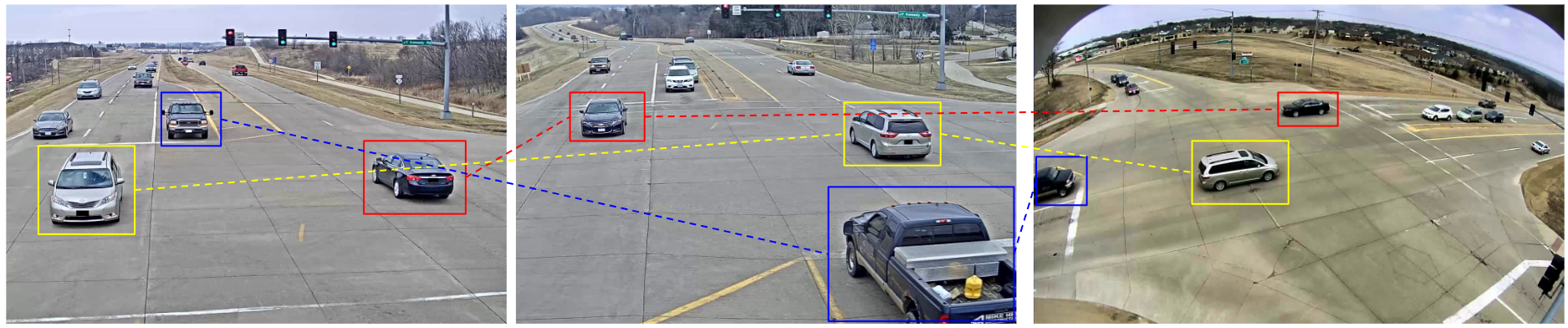}
\caption{Overview of the MTMC task, where different vehicle trajectories are to be captured and matched along many cameras.}
\label{fig:MTMC_example}
\end{center}
\end{figure*}

%%%%%%%%% ABSTRACT
\begin{abstract}
We present TrackNet, a method for Multi-Target Multi-Camera (MTMC) vehicle tracking from traffic video sequences. Cross-camera vehicle tracking has proved to be a challenging task due to perspective, scale and speed variance, as well occlusions and noise conditions. Our method is based on a modular approach that first detects vehicles frame-by-frame using Faster R-CNN, then tracks detections through single camera using Kalman filter, and finally matches tracks by a triplet metric learning strategy. We conduct experiments on TrackNet within the AI City Challenge framework, and present competitive IDF1 results of 0.4733. Our implementation is available at: \href{https://github.com/davidserra9/tracknet}{https://github.com/davidserra9/tracknet}.
\end{abstract}

%%%%%%%%% BODY TEXT
\section{Introduction}
\label{sec:intro}

The continuous growth of urban mobility in cities arises big challenges in the management and optimization of traffic flow. Thanks to the high availability of surveillance cameras and the annotated datasets of road traffic, the task of Multi-Target Multi-Camera vehicle tracking (MTMC) has gained momentum in the Computer Vision community. This task aims to track and identify vehicles along large traffic areas and different road intersections through different video cameras. Good solutions for this problem would allow traffic engineers to improve road design and traffic flow.

The task of MTMC has been always tackled in a multi-stage approach. A first stage is devoted to detect vehicles using object detection algorithms. The second stage performs a tracking of the detected objects in a single video sequence. Finally, the re-identification stage aims to identify the cars along all cameras.

In this paper we propose \textit{TrackNet}, a modular method for multi-camera vehicle tracking based on Faster R-CNN for vehicle detection, Kalman filter for single-camera tracking, and triplet metric learning for cross-camera identification through learned feature similarity.

%-------------------------------------------------------------------------
\section{Related work}
In this section we review methods in the recent literature for the different stages of our method.
\subsection{Object Detection}
Object detection is one of the classic and fundamental computer vision tasks, which given an input image the methods try to produce bounding boxes and classify each object. Two-stage methods handle region proposal and classification separately. For example, Mask R-CNN \cite{maskrcnn} and Faster R-CNN \cite{fasterrcnn} employ Region Proposal Network (RPN) and an additional regressor to suggest bounding boxes and classify the objects inside the proposals. On the other hand, one-stage detectors handle the two tasks simultaneously making them much simpler and faster such as YOLO \cite{yolo} or RetinaNet \cite{retinanet}.

\subsection{Single camera tracking}
The goal of MTSC (Multi-target single-camera tracking) is to estimate the trajectory of the car through a sequence of frames. The majority of the methods follow the same procedure, solving the problem in two steps: first the compute the detections and then perform the tracking. For example, K. Shim \textit{et al.} is an example where a detector (Mask R-CNN) and feature extractor to extract the characteristics of the objects (ResNeXt-50) is used \cite{shim2021multi}. Besides, the tracking is computed using Kalman filter and Hungarian Algorithm.

\subsection{Multi camera tracking}
Multi-camera tracking is a fundamental technique for traffic optimization. Most methods attempt to solve this challenge by first generating trajectories of the detected objects for each camera and then relating them between devices. For example, C. Liu \textit{et al.}, the winners of track 3 of the 5th AI City Challenge \cite{5thcitychallenge}, use a Direction Based Temporal Mask (DBTM) to match the similar vehicles once the tracking in every single camera has been produced. On the other hand, Ristani and Tomasi \cite{mtmcfeatures} proposed a method that did not require any manual annotation beyond the bounding boxes and ids using an adaptative weighted triplet loss to train a feature extraction network. For multi-camera tracking on vehicles the CityFlow \cite{cityflow} dataset has provided an important benchmark from city-scale traffic cameras.

\section{Method}
\subsection{Vehicle detection}
Our method follows the tracking-by-detection paradigm, similar to other state-of-the-art MTSC methods. That is, using an object detector to obtain bounding box locations of vehicles, and later computing the individual trajectories of the objects. Vehicle detection is the basis of both single and multi-camera tracking. Therefore, its effectiveness and reliability directly affects the performance of the entire system. 

In this work, we evaluate three state-of-the-art detection algorithms: Mask R-CNN, RetinaNet and Faster R-CNN. In both of the three approaches we introduce raw input image frames that are processed by a Resnet50 backbone with Feature Pyramid Networks (FPN). The output is a set of bounding boxes and classes that correspond to the objects detected. Further filtering of vehicles is done in order to ignore classes of no interest. 

\subsection{Single-camera Tracking}
Single camera tracking of objects follows three steps. First we ignore regions of the image that are not considered in the dataset. Second, we define an tracking algorithm to assign individual trajectories. Finally, we post-process the tracking by removing low-variant samples, such as parked cars.

\textbf{Region of interest (RoI) filtering:} In some cameras some cars appear outside the Region of Interest (ROI). To remove those detections, the Bounding Boxes are filtered by a camera mask provided in the Dataset (concretely the filtering is used under a distance threshold from the ROI border).

\textbf{Vehicle tracking:} Following the tracking-by-detection paradigm, the bounding boxes at each frame are assigned with a track id corresponding to a single trajectory. We explore two strategies, Kalman Filter and Maximum Overlap. The Maximum Overlap strategy assigns an existing track ID to a detection it overlaps by a $IoU_{0.5}$ threshold with a detection from the track in the past frame. If a detection does not overlap with any previous bounding boxes, a new track ID is assigned. Kalman filter works by propagating an object state into future frames using a linear constant velocity model, independent of other objects and camera motion.

We make use of SORT \cite{bewley2016simple} and DeepSORT \cite{wojke2017simple} implementations of the Kalman filter and the Hungarian Algorithm. In SORT, if a detection is associated to a target, the detected bounding box is used to update the target state where the velocity components are solved optimally by a Kalman filter. If no detection is associated to the target, linear velocity model is used. SORT handles correctly short-term occlusions, but is more unstable for long-term. In turn, DeepSORT integrates appearance information (spatial features) to SORT. The result is the combination, as a weighted sum, of the locations based on motion, which is useful for short term predictions, and appearance, which handles long term occlusions. 

\textbf{Variance filtering:} In some cases parked cars are detected and assigned a track, although our task only considers cars that are in movement. To remove this samples, we come up with a way to detect tracks that are not moving. To that end, the centroid variance of each track is computed and if it is below a variance threshold, in the next frames it is removed. 

\subsection{Cross-camera tracking}
We pose the task of identifying vehicles in different video sequences as a Metric Leaning problem, where we assume that images of the same car will generate similar spatial features. Learned appearance features are projected to a low dimensional space where simiralities can be imposed through a Triplet loss and identifiable clusters can be generated. The triplet loss is defined as,

\begin{equation}
    L_{tri} = \sum_i^N \left[||f(x_i^a) - f(x_i^p)||_2^2 - ||f(x_i^a) - f(x_i^n)|| + \alpha\right],
\end{equation}
where $f(x)$ is the inference function over the learned model, $a$, $p$, and $n$ superscripts stand for anchor, positive and negative samples, and $\alpha$ is the margin. The margin allows to model the minimum distance allowed for negative pairs.

The target of the triplet is to generate representative features for each car, being distinctive between different vehicles and as similar as possible between the same one, even if the vehicle is taken from different cameras. However, the main drawback is that some detections are small because the car is far away from the capture point and the network has troubles in encapsulating the information. Therefore, in order to retain only the richest information from each single-camera track, each vehicle is sampled isochronously obtaining only five detections from the 30\% part of the biggest bounding boxes.

The process of multi-camera tracklets re-ID is done sequentially by pairs, with the set of already re-ided tracks as first input and the corresponding camera as second. The re-id block matches two tracks if they are the most similar ones between them and they fulfill the cross-match condition. This condition ensures that a successful match i.e. the i-th track of camera A has the j-th in set B as the best match and vice-versa. By doing so, we make the system more robust against false matches.

\section{Experimental results}
\subsection{Dataset and evaluation}
The dataset used for both training and testing the proposed method is \textit{CityFlowV2}, the one proposed for the AI City Challenge. This dataset is composed of video sequences acquired by 46 cameras in 16 different city intersections. There is a total of 313931 annotated bounding boxes for 880 distinct vehicles. The original dataset contains 6 different scenarios, 3 for training, 2 for validation and 1 for test. In our setting, we use only three sequences (1, 3 and 4), that are split in a two-fold cross-validation fashion. It shows challenging conditions such as parked cars, occlusions, or lack of bounding boxes when vehicles are at high distance. 

To evaluate the method, the official challenge metrics are considered, namely IDF1, IDP, IDP, Detection Precision and Detection Recall, as described in \cite{MTMC_metric}. The principal evaluation metric for MTMC is IDF1. We also make use of Average Precision (AP) to asses vehicle detectors.

\subsection{Experiment setting}
The proposed method is implemented in Pytorch 1.8.1, with the use of available libraries such as Detectron2 for Object Detection, SORT and DeepSORT for object tracking. We trained the models using a single NVIDIA GeForce RTX 3060 during 5000 epochs. We use AdaGrad optimizer with LR of 0.001 and batch size of 16. 

\textbf{Experiments on vehicle detection:} Table \ref{table:car_detection} presents $AP_{0.5}$ results on different architectures for vehicle detection. We explored RetinaNet, Mask R-CNN and Faster R-CNN with ResNeXt101 backbones, using the pre-trained weights on the COCO dataset and fine-tuning with CityFlowV2. The fine-tuning is highly effective for all architectures , and Faster R-CNN outperforms the other methods. Even though time was not reported, we empirically observed that RetinaNet outperforms the other methods in inference speed, due to its one-stage design. This makes it a better choice in an online or production settings.
\begin{table}[t!]
    \centering
        \begin{tabular}{cc}
            \toprule
                Method & $AP_{0.5}$\\ 
            \midrule
                Off-the-shelf Mask R-CNN & 0.4952\\
                Fine-tuned Mask R-CNN & -\\
                \midrule
                Off-the-shelf RetinaNet & 0.7269\\
                Fine-tuned RetinaNet & 0.9710\\
                \midrule
                Off-the-shelf Faster R-CNN & 0.4813\\
                Fine-tuned Faster R-CNN & 0.9852\\
            \bottomrule
        \end{tabular}
    \caption{Vehicle Detection Results for $AP_{0.5}$, using off-the-shelf ImageNet-pretrained parameters and fine-tuning with CityFlowV2. We do not report Fine-tuned results for Mask R-CNN because CityFlowV2 does not offer masks.}
    \label{table:car_detection}
\end{table}

\textbf{Experiments on Single-Camera Tracking:} Table \ref{table:MTSC} presents IDF1 results and ablation study of the different tracking methods in the single-camera scenario. In all approaches we observe that pre and post processing steps of RoI Filtering (RF) and Variance Filtering (VF) improve results. We conclude that DeepSORT outperforms the others in mostly all cameras.

\begin{table*}[t!]
    \centering
    \begin{tabular}{cccccccccccc}
        \toprule
        & \multicolumn{7}{c}{{Seq.  3}} & \hspace{0.1cm} &  Seq. 1 & \hspace{0.1cm} & Seq. 4\\ 
        \cmidrule{2-8} 
        \cmidrule{10-10} 
        \cmidrule{12-12} 
        Camera id &  c010 & c011 & c012 & c013 & c014 & c015 & \textbf{Avg}  && \textbf{Avg} && \textbf{Avg}\\
        
        \midrule
        Max. Overlap &  0.4210 & 0.3039 & 0.0323 & 0.8781 & 0.3907 & 0.0042 & 0.3384 && 0.7653 && 0.5161\\
        + RF &  0.9239 & 0.3039 &0.0323 &0.8781 &0.6576 & 0.0861 &0.4803&&0.7912&&0.6280\\
        + VF &  0.9446&0.8214&0.2105&0.9092&0.8677&0.1349&\textbf{0.6480}&&0.8334&&0.7439\\
        
        \midrule
        
        SORT &  0.4334 & 0.3003 &0.0327 &0.6833& 0.4010& 0.0042& 0.3229&&0.7864&&0.5338\\
        + RF &  0.9358& 0.3003& 0.0327& 0.7657& 0.7160& 0.0867& 0.4729&&0.8057 &&0.6363\\
        + VF &  0.9401&0.6676&0.1426&0.7657&0.8672&0.1140&0.5692&&0.8057&&0.6642\\
        \midrule
        Deep SORT  & 0.4180&0.3040&0.0324&0.9057&0.3920&0.0042&0.3427&&0.8345&&0.5716\\
        +RF & 0.9292&0.3040&0.0325&0.9057&0.6601&0.0890&0.4867&&0.8594&&0.6881\\
        +VF &0.9327&0.8531&0.0915&0.8914&0.8710&0.1230&0.6271&&\textbf{0.837}8&&\textbf{0.7693}\\
        \bottomrule
        
    \end{tabular}
    \caption{Results on IDF1 metric for the MTSC scenario. Three different settings are presented: First, using Sequence 3 as test and Sequence 1 and 4 for train. Second, using Sequence 1 for test and the other two for train. Finally, using Sequence 4 as test and the other two for train. For each method we show an ablation study, where the baseline is improved using ROI Filtering (RF) and Variance Filtering (VF). We show average IDF1 results for the three settings, and additionally the results for different cameras that recorded Sequence 3.}
    \label{table:MTSC}
\end{table*}

\textbf{Experiments on Multi-camera Tracking:} Table \ref{table:MTMC} presents the IDF1 results for the proposed \textit{TrackNet} method. The results shown in each row correspond to using the sequence i as test and the other two as train. In the last row, the 3-fold cross-validated result is presented, and an IDF1 of 0.4733 is attained, which is competitive with the results of other works. Figure \ref{fig:MTMC_example} shows qualitative results of our method, showing that the re-identification works good, but has some lacks in the bounding box detection due to the Kalman filter refinement. 
\begin{table}[t!]
    \centering
    \begin{tabular}{cccccc}
        \toprule
        Seq. &  IDF1 & IDP & IDR & Prec. & Recall\\ 
        \midrule
        Seq. 1 &  0.5268 & 0.5272 & 0.5262 & 0.8903 & 0.8887\\
        Seq. 3 & 0.4259 & 0.3546 & 0.5330 & 0.6225 & 0.9358\\
        Seq.4 &  0.4672 & 0.5220 & 0.4227 & 0.9298 & 0.7530\\
        \midrule
        Avg &  \textbf{0.4733} & \textbf{0.4679} & \textbf{0.4940} & \textbf{0.8142} & \textbf{0.8592}\\
        \bottomrule
        \end{tabular}
        \caption{Results of the proposed method TrackNet based on DeepSort and Triplet Metric Learning.}
    \label{table:MTMC}
\end{table}
\begin{figure}[htp]
  \centering
  \includegraphics[scale=0.65]{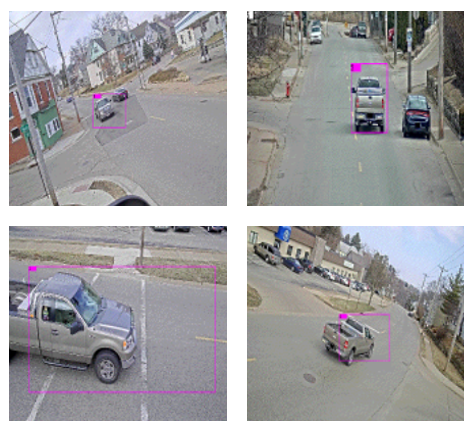}
  \caption{Qualitative results of TrackNet. We observe the tracking and identification of a pickup truck at different cameras and time-steps.}
  \label{fig:MTMC_example}
\end{figure}
\section{Conclusion}
In this work we present \textit{TrackNet}, a method for the task of Multi-Target Multi-Camera Vehicle tracking, and evaluate it on the AI City Challenge framework using CityFlowV2 dataset. Our method is based on Faster R-CNN with Resnet50+FPN for vehicle detection and DeepSORT for tracking. We explore Triplet Metric Learning to learn feature projections that allow for clustering the vehicles across cameras. Our method proves to be effective in the task, showing a IDF1 score of 0.4733, and also giving good qualitative results. The overall performance is sensible to errors in the intermediate layers, hence improvements can be made by trying to enhance the individual stages. The main challenges that we encountered were with occlusions, illumination and noise conditions, high speed vehicles and very similar vehicles. 
{\small
\bibliographystyle{ieee_fullname}
\bibliography{egbib}
}

\end{document}